\def\ANON{0} 
\def\arXiv{1} 

\documentclass[a4paper,twoside]{article}

\usepackage{epsfig}
\usepackage{graphicx}
\usepackage{subcaption}
\usepackage{calc}
\usepackage{amssymb}
\usepackage{amstext}
\usepackage{amsmath}
\usepackage{amsthm}
\usepackage{multicol}
\usepackage{pslatex}
\usepackage{apalike}
\usepackage{flushend}

\usepackage{fancyhdr} 

\usepackage{SCITEPRESS}     

\ifnum\arXiv=1
\fi

\usepackage{xspace}
\usepackage{xcolor}
\definecolor{orange9}{HTML}{FFDD00}
\usepackage[color=orange9]{todonotes}

\begin{document}

\title{Substitution of the Fittest: A Novel Approach for Mitigating Disengagement in Coevolutionary Genetic Algorithms}

\ifnum\ANON=1
    \author{Anon
    \affiliation{}
    \email{}
    }
\else
    \author{\authorname{Hugo Alcaraz-Herrera\sup{1}\orcidAuthor{0000-0002-9991-662X}, John Cartlidge\sup{1}\orcidAuthor{0000-0002-3143-6355}}
    \affiliation{\sup{1}University of Bristol, Bristol, United Kingdom}
    \email{\{h.alcarazherrera, john.cartlidge\}@bristol.ac.uk}
    }
\fi


\abstract{We propose substitution of the fittest (SF), a novel technique designed to counteract the problem of disengagement in two-population competitive coevolutionary genetic algorithms.
The approach presented is domain-independent and requires no calibration. 
In a minimal domain, we perform a controlled evaluation of the ability to maintain engagement and the capacity to discover optimal solutions. 
Results demonstrate that the solution discovery performance of SF is comparable with other techniques in the literature, while SF also offers benefits including a greater ability to maintain engagement and a much simpler mechanism. 
}

\onecolumn \maketitle \normalsize \setcounter{footnote}{0} \vfill



\section{\uppercase{Introduction}}\label{sec:introduction}

While attempting the problem of designing optimal sorting networks using a genetic algorithm (GA), Hillis decided that rather than use randomly generated input lists to evaluate sorting networks, he would instead co-evolve a population of input lists that are evaluated on their ability to not be sorted \cite{hillis90}.
By coupling the evolution of input lists with the evolution of the networks to sort those lists, Hillis attempted to create an ``arms race'' dynamic such that input lists consistently challenge networks that are sorting them. 
As networks improve their ability to sort, so lists become more difficult to sort, etc.
This {\em coevolutionary} approach significantly improved results and generated wide interest amongst evolutionary computation (EC) practitioners.
In particular, coevolution offers EC the ability to tackle domains where an evaluation function is unknown or difficult to operationally define; and through self-learning, coevolutionary systems offer potential for the ``holy grail'' of open-ended evolutionary progress. 

However, it soon emerged that coevolution can suffer from some ``pathologies'' that cause the system to behave in an unwanted manner, and prevent continual progress towards some desired goal. 
For instance, coevolving populations may  continually cycle with no overall progress; populations may progress in an unintended and unwanted direction; or populations may {\em disengage} and stop progressing entirely \cite{Watson2001}. 
These pathologies have been studied in depth and a variety of techniques have been introduced as remedy \cite{Popovici2012}.  
However, there is still much to be understood, and no panacea has been discovered. 

{\bf Contribution:} We propose {\em substitution of the fittest} (SF), a novel domain-independent method designed to tackle the problem of disengagement in two-population competitive coevolutionary systems. 
We explore and evaluate SF in the deliberately simple ``greater than'' domain, specifically designed for elaborating the dynamics of coevolution \cite{Watson2001}.
We compare performance and system dynamics against {\em autonomous virulence adaptation} (AVA), a technique that has been shown to reduce the likelihood of disengagement and improve optimisation in various domains \cite{Cartlidge11}.
Initial results suggest that SF has some benefits over AVA. We evaluate and discuss the reasons why, and present avenues for future investigation.


\section{\uppercase{Background}}\label{sec:related}

Coevolutionary genetic algorithms with two distinct populations are often described using terminology that follows the biological literature. 
As such, and following Hillis' original formulation, the populations are often named as ``hosts'' and ``parasites'' \cite{hillis90}. 
In such cases, the host population tends to denote the population of candidate ``solutions'' that we are interested in optimising (e.g., the sorting networks), while the parasite population tends to denote the population of test ``problems'' for the solution population to solve (e.g., the lists to sort); i.e., the hosts are the {\em models} and the parasites are the {\em training set}; or alternatively the hosts are the {\em learners} and the parasites are the {\em teachers}. 
Throughout this paper, we tend to use the host-parasite terminology to distinguish coevolving populations. 
However, while this terminology is meaningful in {\em asymmetric} systems where one population (the model) is of most interest, it should be noted that in {\em symmetric} systems, such as games of self-play where both coevolving populations are models with the same encoding scheme, the two populations become interchangeable and the names {\em host} and {\em parasite} have less meaning.   

In an ideal scenario, two-population competitive coevolution will result in an {\em arms race} such that both populations continually evolve beneficial adaptions capable of outperforming competitors. 
As a result, there is continual system progress towards some desired optimum goal. 
However, this ideal scenario rarely materialises. 
In practice, coevolutionary systems tend to exhibit pathologies that restrict progress \cite{Watson2001}. 
These include {\em cycling}, where populations evolve through repeated trajectories like players in an endless game of {\em rock-paper-scissors}; and while short-term evolution exhibits continual progress, there is no long-term global progress  \cite{CarBulCIAO-04}.
Alternatively, populations may start to {\em overspecialise} on sub-dimensions of the game, such that evolved solutions are brittle and do not generalise \cite{CarBulECAL03}.
Furthermore, one population may begin to dominate the other to such an extent that populations {\em disengage} and evolutionary progress fails altogether, with populations left to drift aimlessly \cite{Cartlidge04}. 
The likelihood of suffering from these pathologies can be exacerbated by the problem domain. 
Cycling is more likely when the problem exhibits {\em intransitivity}; overspecialisation is more likely in {\em multi-objective} problems; and disengagement is more likely if the problem has an asymmetric {\em bias} that favours one population \cite{Watson2001}. 

Numerous techniques have been proposed for mitigating the pathologies that prevent continual coevolutionary progress (for detailed reviews, see  \cite{Popovici2012,MigCoeTEC18}). 
We can roughly group these approaches into three broad categories; although in practice many techniques straddle more than one category.

First, there are {\em archive} methods, which are designed to preserve potentially valuable adaptations from being ``lost'' during the evolutionary process. 
The first coevolutionary archiving technique is the {\em Hall of Fame} (HoF) \cite{RosinBelew97-HOF}. Every generation, the elite member of each population is stored in the HoF archive. 
Then, individuals in the current population are evaluated against current competitors and also against members of the HoF. 
This ensures that later generations are evaluated on their capacity to beat earlier generations as well as their contemporaries. 
However, as the archive grows each generation, simple archiving methods like the HoF can become unwieldy over time. 
To counter this, more sophisticated and efficient archiving methods have been introduced to simultaneously minimise archive size while maximising  archive ``usefulness''. 
An efficient example is the Layered Pareto Coevolutionary Archive (LAPCA), which only stores individuals that are {\em non-dominated} and {\em unique}; while the archive itself is pruned over time to keep the size within manageable bounds \cite{DeJongLAPCA-07}. 
More recent variations on Pareto archiving approaches include rIPCA, which has been applied to the problem of network security through the coevolution of adversarial network attack and defence dynamics \cite{RIVALS-2017}.
Pareto dominance has also been  employed for selection without the use of an archive, for example the Population-based Pareto Hill Climber \cite{Bar-etal-2018}; and Pareto fronts have been incorporated into an ``extended elitism'' framework, where offspring are selected only if they Pareto dominate parents when evaluated against the {\em same} opponents \cite{akinola-wineberg-2020}. 

A second popular class of approaches attempt to maintain a diverse set of evolutionary challenges through the use of {\em spatial embedding} and {\em multiple populations}. 
Spatially embedded algorithms -- where populations exist on an n-dimensional plane and individuals only interact with other individuals in the local neighbourhood -- have been shown to succeed where other non-spatial coevolutionary approaches fail. 
Explanations for how spatial models can help combat disengagement through challenge diversity have been explored in several works \cite{WieSar-SPATIAL04,WilMitGEC2005}.
Challenge diversity can also be maintained through the use of multiple genetically-distinct populations (i.e., with no interbreeding or migration). Examples include the \emph{friendly competitor}, where two {\em model} populations (one ``friendly'' and one ``hostile'') are coevolved against one {\em test} population \cite{Ficici98a}.
Tests are rewarded if they are both easy to be defeated by a friendly model and hard to be beaten by a hostile model; thereby ensuring pressure on tests to evolve at a challenge-level consistent with the ability of models.
Recently, a new method incorporating the periodic spawning of sub-populations, and then re-integration of individuals that perform well across multiple sub-populations back into the main population has been shown to encourage continual progress in predator-prey robot coevolution \cite{SimNolAlife21}.

Finally, there are approaches that focus on adapting the {\em mechanism for selection} such that individuals are not selected in direct proportion to the number of competitions that they win; i.e., selection favours individuals that are {\em not} unbeatable. 
An early endeavour in this area is the {\em phantom parasite}, which marginally reduces the fitness of an unbeatable competitor, while all other fitness values remain unchanged 
\cite{Rosin97}. Later, the $\Phi$ function was introduced for the density classification task to coevolve cellular automata rules that classify the density of an initial condition \cite{Pagie02}. 
The $\Phi$ function translates all fitness values such that individuals are rewarded most highly for being equally difficult and easy to classify (i.e., by being classified correctly half of the time); while individuals that are always classified or always unclassified are punished with low fitness. 
However, while $\Phi$ worked well, it was limited by being domain-specific. 

More generally applicable is the {\em reduced virulence} technique \cite{CarBul-CEC02,Cartlidge04}. 
Inspired by the behaviour of biological host-parasite systems, where the virulence of pathogens evolves over time, reduced virulence is the first domain-independent technique with tunable parameters that can be configured. 
After generating a parasite score through competition, reduced virulence applies the following non-linear function to generate a fitness for selection:
\begin{equation}\label{eq:rv}
    f(x_i,\upsilon) = \frac{2x_i}{\upsilon} - \frac{x_i^2}{\upsilon^2}
\end{equation}
where $0 \leq x_i \leq 1$ is the relative (or subjective) aptitude of individual $i$ and $0.5 \leq \upsilon \leq 1$ represents the virulence of the parasite population. When $\upsilon=1$, equation~(\ref{eq:rv}) preserves the original ranking of parasites (i.e., the ranking of competitive score, $x$) and is equivalent to the canonical method of rewarding parasites for all victories over hosts. 
When $\upsilon=0.5$, equation~(\ref{eq:rv}) rewards maximum fitness to parasites that win exactly half of all competitions. 
Therefore, in domains where there is a bias in favour of one population (the ``parasites''), setting a value of $\upsilon<1$ for the advantaged population reduces the bias differential in order to preserve coevolutionary engagement. 
Reduced virulence demonstrated improved performance, but is limited by requiring $\upsilon$ to be determined in advance. 
In many domains, bias may be difficult to determine and may change over time. 
To tackle this problem, reduced virulence has been incorporated into a human-in-the-loop system enabling a human controller to {\em steer} coevolution during runtime by observing the system behaviour and altering the value of $\upsilon$ in real time \cite{BulCarThoESTA02}.

Later, {\em autonomous virulence adaptation} (AVA) -- a machine learning approach that automatically updates $\upsilon$ during coevolution -- was proposed \cite{Cartlidge11}. Each generation $t$, AVA updates $\upsilon$ using:
\begin{equation}\label{eq:ava_v_t_plus_one}
    \upsilon_{t+1} = \upsilon_{t} + \Delta_t
\end{equation}

\begin{equation}\label{eq:ava_delta_t}
    \Delta_t = \mu\Delta_{t-1} + \alpha(1 - \mu)(\tau - \overline{X_t})
\end{equation}
where $0 \leq \alpha, \mu, \tau \leq 1$ are learning rate, momentum, and target value, respectively; and $\overline{X_t}$ is the normalised mean subjective score of the population.\footnote{For the initial $t<5$ generations, to avoid immediate disengagement in cases of extreme bias differential, equation~(\ref{eq:ava_delta_t}) is replaced by $\Delta_t = (0.5-\overline{X_t})/t$; so virulence can immediately adapt to high ($\upsilon=1$) or low ($\upsilon=0.5$) values.} 
Rigorous calibration of AVA settings demonstrated that values $\alpha = 0.0125$, $\mu = 0.3$, and $\tau = 0.56$ can be applied successfully in a number of diverse domains. 
In particular, it was shown that AVA can coevolve high performing sorting networks and maze navigation agents with much greater computational efficiency than archive techniques such as LAPCA \cite{Cartlidge11}.

\section{\uppercase{Substitution of the Fittest}}\label{sec:sf}

We introduce substitution of the fittest (SF), a novel technique designed to combat disengagement that is domain-independent and requires no calibration. Disengagement tends to occur when one population ``breaks clear'' of the competing population such that all individuals in the leading population outperform all individuals in the trailing population. Therefore, in simple terms, SF is designed to apply a ``brake'' to the population evolving more quickly; while for the population trying to keep pace, SF applies an ``acceleration''. Consequently, the advantage of the leading population over the trailing population is reduced. In this way, SF is designed to keep populations engaged.

Unlike standard evolutionary algorithms, where individuals are evaluated using an ``absolute'' fitness function that is exogenous to the evolutionary process, competitive coevolutionary GAs utilise a ``relative'' (or ``subjective'') fitness evaluation, where fitness $\psi_i$ of an individual $i$ is endogenously assigned based on performance against other evolving individuals. Usually, score $\psi_i$ is simply the proportion of ``victories'' that $i$ secures across a series of competitions against evolving opponents. 
These competitive interactions between coevolving populations describe a coupled system that has potential to develop into an arms-race of continual progress. 
However, when the populations decouple, i.e., when disengagement occurs, all information regarding the relative differences in performance of individuals is lost, such that $\forall i,j: \psi_i=\psi_j$. 
This is problematic and causes the coevolving populations to drift.

The current state of a population can be measured by the {\em population mean subjective aptitude}: 
\begin{equation}\label{eq:mean_subjective_aptitude}
    \sigma_{pop} = \frac{\sum_{i=1}^{n}\psi_{i}}{n}
\end{equation}
where $n$ is the number of individuals in the population and $0 \leq \sigma \leq 1$. Then, $\sigma$ values for each population can be used to measure the {\em level of disengagement}, defined as: 
\begin{equation}\label{eq:disengagement_value}
    \delta = |\sigma_{popA} - \sigma_{popB}|
\end{equation}
 where $0 \leq \delta \leq 1$. When populations have similar $\sigma$, then disengagement $\delta$ has a low value close to zero. When populations are fully disengaged, i.e., when $\sigma_{popA}=1$ and $\sigma_{popB}=0$, or when $\sigma_{popA}=0$ and $\sigma_{popB}=1$, then $\delta=1$.
 During the coevolutionary process, $\delta_t$ is calculated and stored each generation $t$. If $\delta_{t+1}\leq\delta_t$, disengagement {\em did not} increase; otherwise, disengagement {\em did} increase and so SF is triggered. 
 
 When SF is triggered, we first calculate the number of individuals to be substituted, $\kappa$, defined as:
\begin{equation}\label{eq:number_of_clones}
    \kappa = n\delta^{\frac{1}{\delta}}
\end{equation}
 where $n$ is the number of individuals in the population and the result is rounded up to the nearest integer. 
 The value of $\kappa$ increases non-linearly as a function of $\delta$.
 As populations approach disengagement ($\delta$ near 1), the number of substitutions tends to $n$. 
 It is important to point out that if $\kappa>\frac{n}{2}$, then effectively only $n-\kappa$ substitutions occur. 
 For instance, given a population with $n=6$ parasites whose aptitudes are [0.8,0.6,0.4,0.2,0.1,0.0] and $\kappa=4$, after the substitution, their aptitudes will be [0.0,0.1,0.2,0.4,0.1,0.0]. 
 In this example, half the population was not modified and the two fittest individuals were substituted by the two worst individuals. 
 Furthermore, if $\delta=1$, populations would not be modified as the number of individuals to be substituted is the same as the population size  (i.e., $\kappa=n$).   
On the contrary, when $\delta<0.3$, $\kappa$ tends to 0. 
During these times,  populations are sufficiently engaged and substitutions are not necessary. 
 
 The next step in the SF process consists of comparing $\sigma_{popA}$ and $\sigma_{popB}$ and then substituting $\kappa$ individuals in each population using the following rules:

\begin{itemize}
    \item \emph{Population with lowest $\sigma$}: 
    Rank all individuals by subjective aptitude $\psi_i$. Then, the $\kappa$ individuals with the {\em highest} $\psi_i$ replace the $\kappa$ individuals with the {\em lowest} $\psi_i$. Finally, the subjective aptitude of every individual is increased by:
    \begin{equation}\label{eq:host_aptitude}
     \psi_{i}'= min(\psi_{i} + \delta,1)
    \end{equation}
    taking minimum value to ensure $0 \leq \psi_{i}' \leq 1$.
    \item \emph{Population with highest $\sigma$}: Rank all individuals by $\psi_i$. Then $\kappa$ individuals with the {\em lowest} $\psi_i$ replace the $\kappa$ individuals with the {\em highest} $\psi_i$; following replacement, the subjective aptitude of every individual is decreased using:
    \begin{equation}\label{eq:parasite_aptitude}
     \psi_{i}'= max(\psi_{i} - \delta,0)
    \end{equation}
    taking maximum value to ensure $0 \leq \psi_{i}' \leq 1$.
\end{itemize}

As described, SF affects each population in a different manner.
For the population that evolves more quickly, the proportion of individuals to be randomly selected is increased. 
For instance, if the highest ranked individual whose $\psi<\delta$, then the $\kappa$ individuals would be selected at random because they would have $\psi'=0$. 
On the other hand, for the population which evolves slower than the other, the effect is the opposite, i.e., if the lowest ranked individual whose $\psi+\delta\geq 1$, then all $\kappa$ individuals would have $\psi'=1$ and hence those individuals would have high probability of being selected.  

Following SF, selection is performed and genetic operators are applied as usual.
In the minimal experiments we present in the following sections, we use tournament selection and apply mutation, i.e., populations are asexual and recombination is not used.


\section{\uppercase{Experimental Method}}\label{sec:method}

\subsection{The ``greater than'' game}\label{sec:counting_ones}

\begin{figure*}[tb!]
  \centering
     \includegraphics[width=0.4\linewidth]{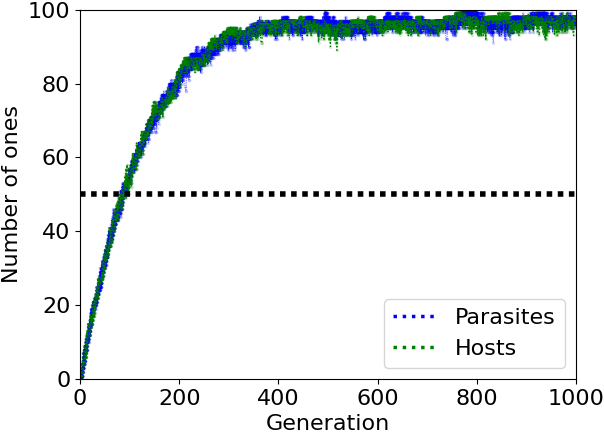}
     \includegraphics[width=0.4\linewidth]{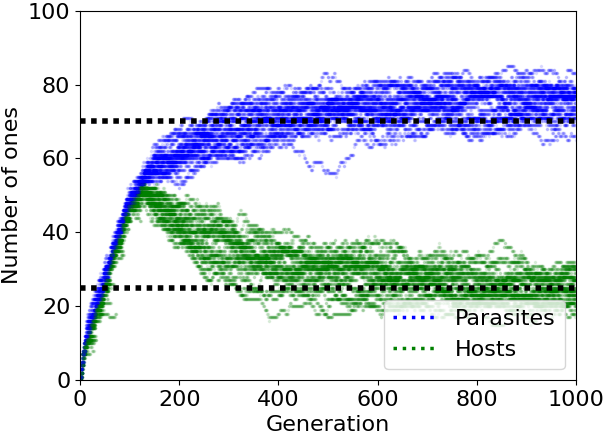}%
  \caption{Coevolution: (left) equal bias $\beta_h = \beta_p = 0.5$; and (right) differential bias $\beta_h = 0.25$, $\beta_p = 0.75$.}
  \label{fig:baseline}
\end{figure*}

The \emph{greater than} game \cite{Watson2001} was introduced as a minimal (and analytically tractable) substrate capable of demonstrating some of the pathological dynamics of coevolution; in particular {\em disengagement}. 
The game consists of maximising scalar values through a comparison-based function where given two scalar values, $\alpha$ and $\gamma$, the function operates as $score(\alpha,\gamma)=1$ if $\alpha>\gamma$, 0 otherwise.
Here, we use a slightly modified ``greater than or equals'' game that rewards draws; such that $score(\alpha,\gamma)=1$ if $\alpha>\gamma$; $0.5$ if $\alpha=\gamma$; 0 otherwise.

The coevolutionary set-up consists of two isolated populations, each with $n$ individuals. 
Each individual is represented by a binary string with $l=100$ bits
and the ``objective'' purpose of the coevolutionary system is to evolve individuals with bit-strings containing all ones (i.e., scalar values of $l=100$).
Every generation, to generate a subjective aptitude score, each individual plays the greater than game against a sample of $S$ opponents. 
Tournament selection is used to select individuals for reproduction, and the only genetic operator is mutation (i.e., reproduction is {\em asexual}). 
Mutation has a bias controlled by parameter $\beta$, where $0\leq\beta\leq1$. 
For each bit, there is a probability $m$ of mutation occurring. 
When it occurs, the bit is assigned a new value at random, with probability $\beta$ of assigning a 1, and probability $1-\beta$ of assigning a 0; i.e., when $\beta=0.5$, mutation has an equal chance of assigning the bit to 1 or 0; when $\beta=0$ mutation will always assign the bit to 0; and when $\beta=1$ mutation will always assign the bit to 1. 
This bias parameter $\beta$ allows the simple game to emulate the intrinsic asymmetry of real and more complex domains, where it is often easier for one population to outperform another.
Under mutation bias alone, i.e., when populations are disengaged and left to drift under the absence of selection pressure, we expect the population to tend towards having $\beta \times l$ ones. 
Therefore, for a bias $\beta=0.5$, we would expect the population to drift towards scalar values of 50.

In our two-population competitive set up, we label the populations as as {\em hosts} and {\em parasites}. Each population has an independent bias value $\beta$, which controls the problem difficulty for each population.  We use $\beta_h$ to label the bias value of the host population and $\beta_p$ to label the bias value of the parasite population. When bias differential is high, i.e., when the value of $\beta_p$ is much larger than the value of $\beta_h$ (or vice versa), disengagement becomes more likely as the game is much easier for the parasites (alternatively, the hosts) to succeed. In more complex domains, there is often asymmetry in problem difficulty for coevolving populations. By varying $\beta_p$ and $\beta_h$, we are able to control the asymmetry of problems in the simple greater than game.

Our experimental set up is detailed as follows. We coevolve two isolated populations, each with 25 individuals ($n=25$). The length of the binary array (an individual) is 100 ($l=100$) and each bit is initialised to 0. 
For generating a competitive score, we use a sample size of 5 ($S=5$). 
Tournament size for selection is 2. The probability of mutation per bit is 0.005 ($m=0.005$). Finally, each evolutionary run lasts for 1000 generations.

\begin{figure*}[tb!]
  \centering
  \includegraphics[width=0.35\linewidth]{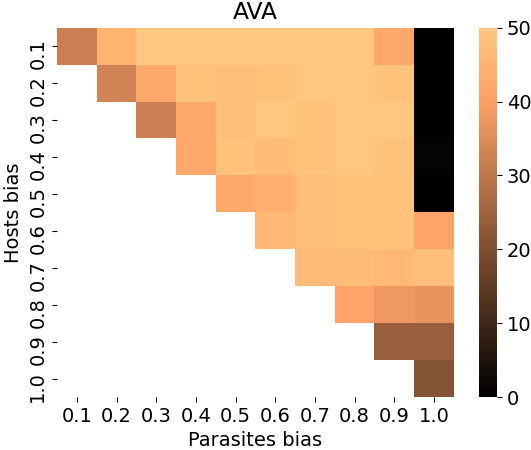}%
  \includegraphics[width=0.35\linewidth]{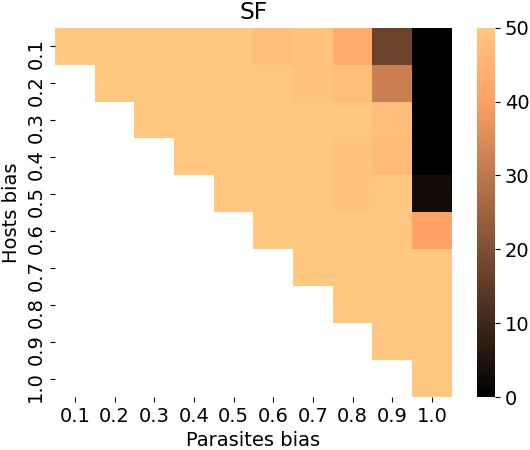}
  \centering
  \caption{Number of runs with no disengagement; AVA (left) and SF (right) across all bias levels (50 trials).}
  \label{fig:heatmaps_engagement}
\end{figure*}

\begin{figure*}[tb!]
  \centering
  \includegraphics[width=0.35\linewidth]{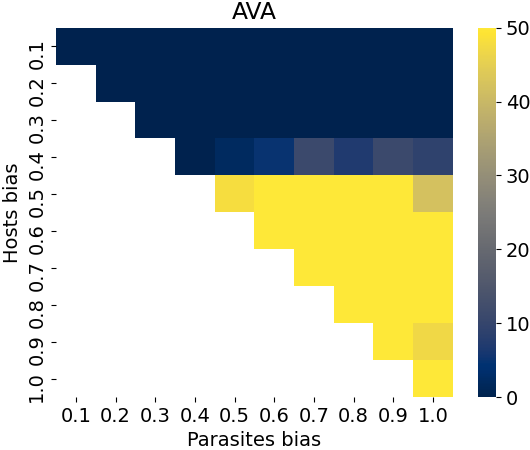}%
  \includegraphics[width=0.35\linewidth]{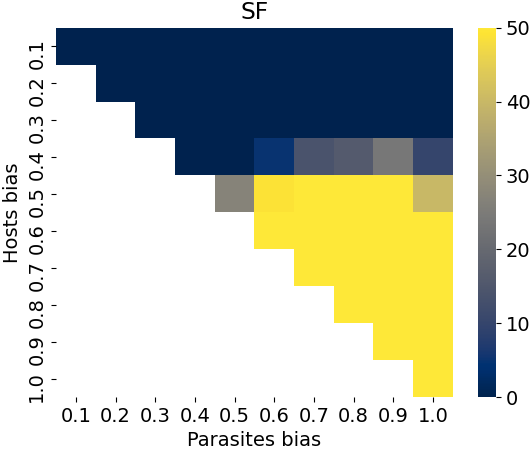}
  \centering
  \caption{Number of runs where hosts reached optimum; AVA (left) and SF (right) across all bias levels (50 trials).}
  \label{fig:heatmaps_reach_100}
\end{figure*}

\begin{figure*}[tb!]
  \centering
  \includegraphics[width=0.35\linewidth]{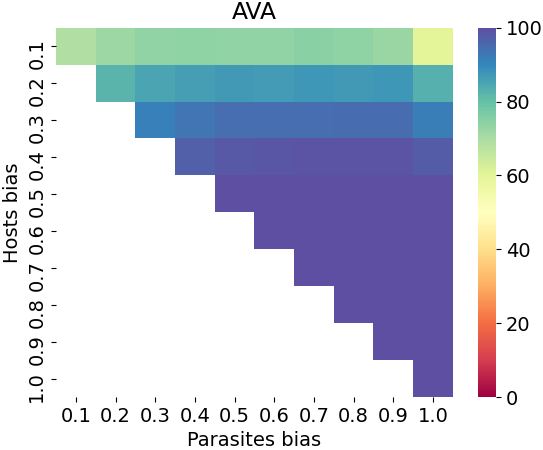}%
  \includegraphics[width=0.35\linewidth]{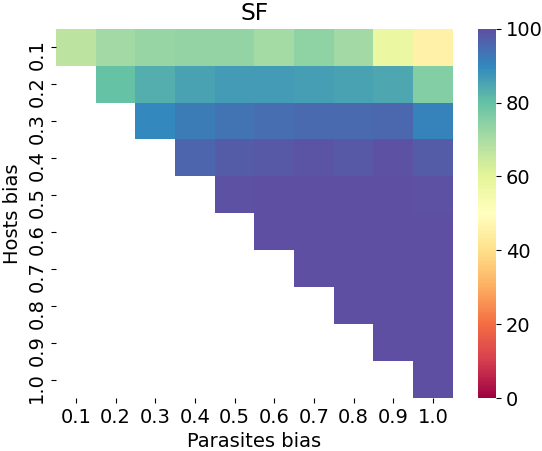}
  \centering
  \caption{Mean number of ones of best host; AVA (left) and SF (right) across all bias levels (50 trials).}
  \label{fig:heatmaps_max_ones}
\end{figure*}

\subsection{Disengagement}\label{sec:standard_cga}

\noindent
The effect of disengagement can most clearly be elaborated by visualising what happens when it occurs. 
Figure~\ref{fig:baseline} (left) presents the  coevolution of two populations, each with equal bias $\beta_p=\beta_h=0.5$.
We see that populations remain engaged throughout the evolutionary run. This engagement provides a continual gradient for selection and encourages an ``arms race'' of increasing performance. As a result, both populations reach optimal performance of 100 ones. This is far higher than 50 ones (dotted line) that both populations would be expected to reach if drifting through space under mutation alone, i.e., when selection pressure is removed. 

In contrast, Figure~\ref{fig:baseline} (right) demonstrates the pathology of disengagement. Here, there is a differential bias in favour of parasites, such that $\beta_p=0.75$ and $\beta_h=0.25$.
Initially, both populations remain engaged and selection drives evolutionary progress, with both populations reaching approximately 50 ones by generation 150. However, the impact of differential bias in favour of parasites then leads to a disengagement event such that all parasites have more ones than their competing hosts; resulting in a subjective score of zero for all hosts and a subjective score of one for all parasites. At this point selection pressure is removed as all individuals have an equal (and therefore random) chance of selection, leaving the populations to drift under mutation alone. As expected, the parasite population drifts to the parasite mutation bias (dotted line) of 75, while hosts degrade to the host mutation bias of 25.
The high bias differential between populations ($\beta_p - \beta_h = 0.5$) not only causes the initial disengagement event, but ensures that post-disengagement populations drift through different regions of genotype/phenotype space and will never re-engage through chance alone. 

In general, the greater the bias differential between populations, the greater the likelihood of disengagement occurring. As shown, disengagement severely hinders coevolutionary progress.

\section{\uppercase{SF vs AVA: A Comparison}}\label{sec:results}

To measure the performance of SF, we perform a thorough comparison against AVA \cite{Cartlidge11}, which has been previously shown to dramatically reduce the effects of disengagement in the greater than game and also in several more complex and realistic domains, including designing minimal-length sorting networks and discovering classifier systems for maze navigation.
To understand how the two approaches are likely to perform in more complex domains, where population asymmetries are more likely, we trial both SF and AVA in simulations where mutation bias is varied across all possible levels  ($\beta_p \geq \beta_h$) in the range [0.1, 1.0]. 
The mutation bias was configured in favour of parasites (except when $\beta_p = \beta_h$); although this decision is arbitrary and results where bias is in favour of hosts would yield symmetrically similar results. 
For each bias scenario, we performed 50 experimental trials. 
To analyse performance of SF and AVA, we utilise three metrics: (i) the reliability of the technique to maintain population engagement; (ii) the capacity to discover optimal hosts containing all ones; and (iii) the mean number of ones that hosts reach before disengagement occurs.
The following sections describe our findings.

\begin{figure*}[tb!]
  \centering
  \includegraphics[width=0.4\linewidth]{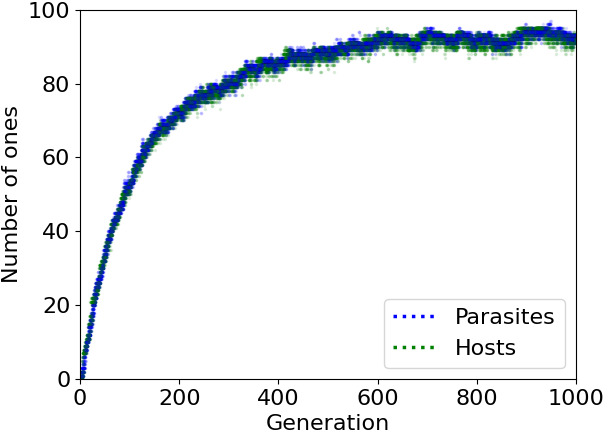}%
  \includegraphics[width=0.4\linewidth]{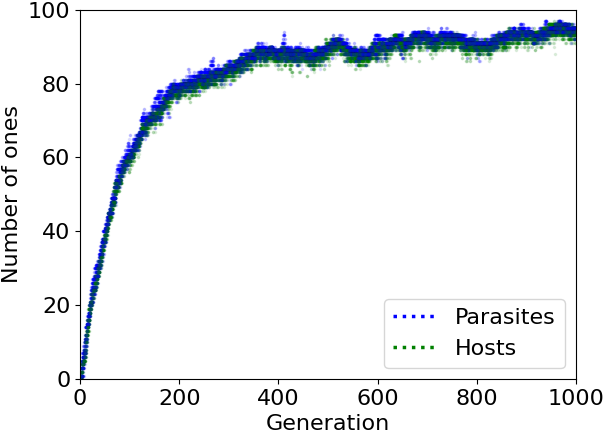}
  \centering
  \caption{Example coevolutionary runs under AVA (left) and SF (right) with $\beta_h = 0.3$, $\beta_p = 0.7$.}
  \label{fig:ava_vs_sf2}
\end{figure*}

\begin{figure*}[tb!]
  \centering
  \includegraphics[width=0.4\linewidth]{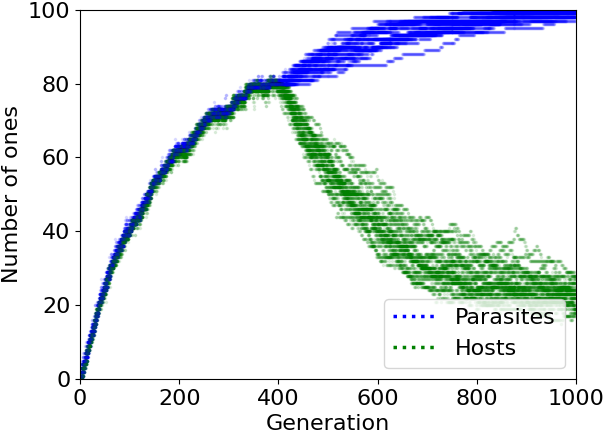}%
  \includegraphics[width=0.4\linewidth]{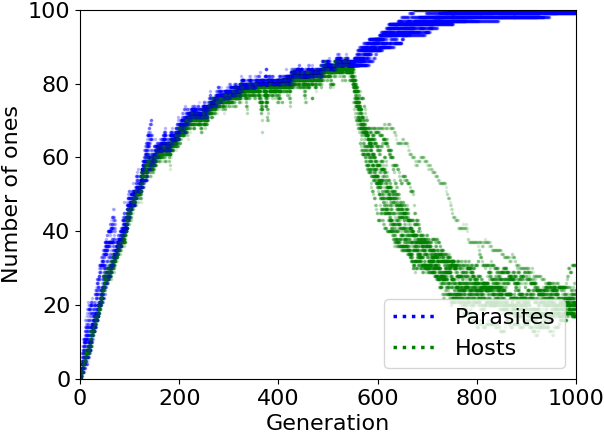}
  \caption{Example coevolutionary runs under AVA (left) and SF (right) with $\beta_h = 0.2$, $\beta_p = 1.0$.}
  \label{fig:ava_vs_sf1}
\end{figure*}

\subsection{Maintaining engagement}\label{sec:reliability}

Since the main objective of AVA and SF is to maintain engagement during coevolution, regardless of bias fluctuations that may arise, it is fundamental to study their response under diverse bias levels. 
Figure \ref{fig:heatmaps_engagement} presents a heatmap showing the number of runs where AVA and SF maintained population engagement during the full coevolutionary process (regardless of whether an optimal host is found).  

Overall, we see that SF is able to maintain engagement more successfully than AVA across a diverse range of bias differentials. For 38 of the 50 bias pairings, SF maintains engagement for the full coevolutionary run of 1000 generations across all 50 trials. For AVA, however, this number is only 12. Moreover, SF maintains engagement for the full coevolutionary run in at least 40 out of 50 trials across 49 bias pairings, while for AVA this number is 42. 
 
Relative to SF, results suggest that AVA tends to struggle in scenarios (i) where the bias of both populations are either the same (symmetrical systems) or similar; and (ii) where parasite bias is very high (e.g., $\beta_p=1.0$) and there is a large bias differential between parasites and hosts. 
In comparison, although SF also fails in scenarios where parasites have very high bias, SF is capable of maintaining engagement where AVA is not.    

Regarding the result obtained by AVA, in the original research \cite{Cartlidge11}, it was calibrated to only handle bias levels in the range [0.5, 1.0]. Moreover, in the original experiments the number of generations is 750 whereas in these experiments, the number of generations is 1000. The duration of experimental trials is a key factor inasmuch AVA, in a number of bias scenarios, tends to allow disengagement {\em after} optimal hosts are found. 
For instance, when $\beta_h = 0.5, \beta_p=1.0$, populations tend to first reach the optimum, but then later, around generations 850-900, the populations disengage. 
This unexpected behaviour suggests that AVA parameters may require  recalibration to maintain engagement over long time periods when there is a high bias differential. 
It also demonstrates an advantage of SF over AVA, as SF has no parameter settings to calibrate.

\subsection{Reaching the optimum}
\noindent
Another essential aspect to analyse is the capability to reach the optimal zone. 
Figure \ref{fig:heatmaps_reach_100} presents the number of runs where hosts (more precisely, at least one host) reached the optimal, regardless of whether or not populations disengage after this point. 
We see a similar pattern for both SF and AVA. 
AVA reached the optimal zone at least 40 times under 21 bias levels, where as SF was capable of reaching 40 times or more under 20 bias levels.
Furthermore, both AVA and SF reached the optimal zone a maximum of 50 times (i.e., every time) under 18 bias scenarios.

As expected, in multiple bias scenarios hosts were not capable of reaching the optimum when hosts have a very low mutation bias ($\beta_h < 0.5$).

\subsection{General performance}\label{sec:performance}

Figure \ref{fig:heatmaps_max_ones} shows the mean maximum number of ones of the best host across all bias configurations, regardless of whether or not populations disengage or hosts reach the optimum. 
Again, performance of SF and AVA is similar. AVA reaches at least 90 ones under 36 bias scenarios, whereas SF reaches 90 ones in 35 scenarios. Furthermore, AVA and SF both reach 100 ones under 18 possible bias scenarios.

Results suggest that AVA and SF tend to behave similarly across most bias levels. 
 However, when there is a significant bias differential  (e.g., $\beta_h=0.1,\beta_p=0.9$), AVA enables populations to reach a greater performance (closer to the optimum) than SF.





\subsection{Coevolutionary dynamics}\label{sec:dynamic_populations}

Figures \ref{fig:ava_vs_sf2} and \ref{fig:ava_vs_sf1} present example runs to highlight the effects of AVA (left) and SF (right) during the coevolutionary process. 
When bias differential is relatively large (Figure~ \ref{fig:ava_vs_sf2}; $\beta_h = 0.3$ and $\beta_b = 0.7$) both AVA and SF maintain engagement throughout. 
However, when bias differential is very large (Figure~\ref{fig:ava_vs_sf1}; $\beta_h = 0.2$ and $\beta_b = 1.0$), disengagement occurs under both techniques, but tends to occur earlier (around generation 410 for AVA, compared with generation 550 for SF). 

Interestingly, in Figure~\ref{fig:ava_vs_sf1} we see that SF induces different population dynamics with respect to AVA. 
Under SF, in the early generations, parasites exhibit lots of variation, with some outliers drifting far from the engaged populations. Then, around generations 90 and 170, these outlier lineages suddenly disappear. 
A similar ``cull'' effect is {\em not} observed with AVA. 

\begin{figure}[tb!]
  \centering
  \includegraphics[width=1.0\linewidth]{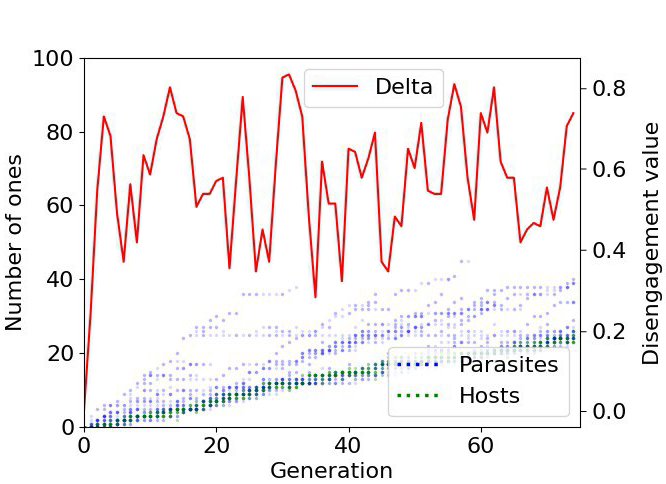}
  \centering
  \caption{Initial generations of one run ($\beta_h=0.1$, $\beta_p=1.0$) using SF, showing absolute fitness of parasites and hosts (left axis) and $\delta$ value of disengagement (right axis).} 
  \label{fig:cull}
\end{figure}

To further investigate the cull effect in SF and its direct relation with disengagement value $\delta$, Figure~\ref{fig:cull} presents the initial generations of one example run with very high bias differential ($\beta_h=0.1$, $\beta_p=1.0$). 
Around generation 30, we see that there is a single outlier parasite containing around 40 ones. 
As a consequence, $\delta\approx0.81$ and $\kappa = 25(0.81^{1/0.81}) = 19$ (see Equation~\ref{eq:number_of_clones}). 
Since $19 > \frac{25}{2}$, the effective substitutions are 6 (see example given in Section \ref{sec:sf}). 
Therefore, the 6 fittest parasites, including the outlier, are substituted by the 6 worst parasites. 
Consequently, the subsequent generations do not present outliers and $\delta$ (and therefore $\kappa$) decreases.
Other outlier lineages later begin to emerge and the process repeats. 
In contrast, AVA tends to keep both populations more tightly coupled throughout.


\section{\uppercase{Conclusions}}\label{sec:conclusion_future}

This research has introduced SF as an alternative technique to mitigate disengagement in competitive coevolutionary genetic algorithms. Using a minimal problem domain to enable exposition, we compared the performance of SF with AVA, a technique in the literature that has been shown to combat disengagement in a variety of domains.  Experimental results suggest that, in general, SF has similar performance to AVA in terms of discovery of optimal solutions. However, SF is also shown to have better performance than AVA in terms of consistently maintaining engagement across a wide variety of bias differentials, i.e., where there is a large inherent advantage in favour of one coevolving population. 
The mechanism of SF is deliberately designed to be simple and domain independent, requiring no domain knowledge or specific calibration.  This makes SF more easy to implement than other techniques and offers the possibility of being more generally applicable. 

However, one of the potential weaknesses observed in SF is the highly-fluctuating behaviour induced in populations (i.e., the ``cull'' effect), which might lead to sudden disengagement in other more realistic domains. 
Thus, we believe that SF deserves further exploration; although it has shown suitable performance in a simple domain,  experiments in more complex domains such as maze navigation or sorting networks \cite{Cartlidge11} are necessary to demonstrate its reliability. Furthermore, a robust comparison against other state-of-art techniques will be performed. 
Finally, we intend to explore the effects that SF has on other coevolutionary pathologies, such as overspecialisation and cycling. 


\section*{ACKNOWLEDGEMENTS}

Hugo Alcaraz-Herrera's PhD is supported by The Mexican Council of Science and Technology (Consejo Nacional de Ciencia y Tecnología - CONACyT). John Cartlidge is sponsored by Refinitiv.


\bibliographystyle{apalike}
{\small
\bibliography{sf_references}}

\end{document}